\pdfoutput=1

\documentclass[11pt]{article}

\usepackage[preprint]{acl}

\usepackage{times}
\usepackage{latexsym}
\usepackage{float}
\usepackage[T1]{fontenc}
\usepackage{tabularx} 
\usepackage{booktabs} 
\usepackage[utf8]{inputenc}
\usepackage{multirow}
\usepackage{microtype}
\usepackage{multirow}
\usepackage{booktabs}

\usepackage{inconsolata}

\usepackage{graphicx}

\newcommand{\Dataset}{{\textsc{RedditESS}}}

\title{\Dataset:A Mental Health Social Support Interaction Dataset \\ \large Understanding Effective Social Support to Refine AI-Driven Support Tools}

\author{
Zeyad Alghamdi\textsuperscript{\footnotemark[1]}\textsuperscript{1,2} ~~ Tharindu Kumarage\thanks{These authors contribute to this work equally.}\textsuperscript{1} ~~ Garima Agrawal\textsuperscript{3} ~~ Mansooreh Karami\textsuperscript{4} \\ ~~\textbf{Ibrahim Almuteb}\textsuperscript{4}~~ \textbf{Huan Liu}\textsuperscript{1} \\
         \textsuperscript{1} Arizona State University, 
         \textsuperscript{2}Hail University, 
         \textsuperscript{3}HumaConn AI Consulting, 
          \textsuperscript{4}Microsoft, 
         \textsuperscript{5}Texas A\&M University \\
         \texttt{\{zalgham1, kskumara, huanliu\}@asu.edu} ~~
         \texttt{garima@humaconn.com} ~~ \\
         \texttt{mkarami@microsoft.com} ~~
         \texttt{iknm@tamu.edu}
         }

\begin{document}
\maketitle
\begin{abstract}

Effective mental health support is crucial for alleviating psychological distress. While large language model (LLM)-based assistants have shown promise in mental health interventions, existing research often defines "effective" support primarily in terms of empathetic acknowledgments, overlooking other essential dimensions such as informational guidance, community validation, and tangible coping strategies. To address this limitation and better understand what constitutes effective support, we introduce {\Dataset}, a novel real-world dataset derived from Reddit posts, including supportive comments and original posters' follow-up responses. Grounded in established social science theories, we develop an ensemble labeling mechanism to annotate supportive comments as  effective or not and perform qualitative assessments to ensure the reliability of the annotations. Additionally, we demonstrate the practical utility of {\Dataset} by using it to guide LLM alignment toward generating more context-sensitive and genuinely helpful supportive responses. By broadening the understanding of effective support, our study paves the way for advanced AI-driven mental health interventions. Our dataset is available at the following \href{https://anonymous.4open.science/r/RedditESS-3577}{repository.}

\end{abstract}

\section{Introduction}

Social support encompasses the provision of emotional, informational, and instrumental resources designed to help individuals navigate stressful life events and mental health challenges~\cite{house1988structures,yang2023information}. Effective social support can mitigate psychological distress, enhance resilience, and improve overall well-being~\cite{cohen1985stress,rini2011effectiveness}. Within mental health contexts, providing appropriate support is crucial not only for healthcare professionals and peers but increasingly for artificial intelligence (AI) systems~\cite{hua2024large,lawrence2024opportunities}. Large Language Models (LLMs) have demonstrated potential as moderators in online mental health communities, offering supportive and non-judgmental responses that may alleviate isolation, foster understanding, and facilitate positive interactions~\cite{de2023benefits,guo2024large}. Ensuring that these AI-driven agents can deliver consistently effective support holds significant promise for accessible, scalable, and immediate assistance, particularly in digital environments where human support may be limited ~\cite{molli2022effectiveness,almakinah2024enhancing}.

Most existing AI-driven efforts to enhance mental health support have mainly focused on generating empathetic responses~\cite{loh2023harnessing,chen2023soulchat,kearns2024bridging}. Existing datasets, often derived from controlled environments~\cite{sharma2020engagement}, clinical settings~\cite{lai2023supporting, lai2023psy}, or limited interaction types~\cite{medeiros2018using}, have narrowly defined effective support through empathy alone~\cite{sharma2020computational}. While empathy is undoubtedly important, focusing solely on it overlooks other critical attributes of effective support. For example, individuals may value informational guidance, validation, encouragement, or tangible coping strategies just as highly as empathetic acknowledgments ~\cite{shen2024empathy,rubin2024considering}. Moreover, previous datasets frequently lack feedback loops from original posters (OPs), rendering it challenging to assess the perceived quality and impact of provided support accurately~\cite{althoff2016large,perez2015experiments}.

To address these limitations, we introduce a novel dataset, \Dataset, designed to capture multiple dimensions of effective social support in real-world digital settings. Sourced from Reddit, a platform conducive to open and authentic discussions about mental health~\cite{de2014mental,alghamdi2024less}, our dataset consists of original posts describing stressful or distressing situations, subsequent comments offering support, and, most importantly, the OP's replies to these comments. In addition, we collect metadata related to these interactions, including upvotes and controversy scores provided by Reddit. This three-tier interaction structure and accompanying metadata enable a more nuanced approach to evaluating the effectiveness of social support. 

Specifically, here we focus on two primary dimensions inspired by social science and psychological theories while defining effective social support: \textit{reciprocity} and \textit{community reception}. According to~\cite{rini2011effectiveness}, effective support comprises emotional, informational, and instrumental resources perceived as reciprocal, where the recipient actively engages with and responds to the support provider. Building on prior research emphasizing reciprocity as a key indicator \cite{feng2010influences, cutrona1992controllability, rime2009emotion, burleson1996comforting}, we prioritize the original poster's engagement and feedback to evaluate how well the support resonates with the individual's needs. We then incorporate the second dimension of community reception using upvotes, controversy scores, and other crowd-based indicators to reflect `community-validated' supportive responses \cite{andalibi2017sensitive, de2014mental, chancellor2016quantifying}. 
By integrating these dimensions, we establish a holistic and robust labeling of `effective' social support. We further employ LIWC (Linguistic Inquiry and Word Count) to analyze linguistic and affective features associated with supportiveness.
Following the dataset construction, we perform comprehensive qualitative evaluations with human annotators to assess the reliability and clarity of our labeling process. To demonstrate the practical utility of {\Dataset}, we incorporate it into LLM training pipelines through instruction tuning and alignment, treating effective social support comments as human preference data. Our experiments reveal that integrating this data enhances the models' ability to generate effective supportive responses. 

In summary, our key contributions are:
\begin{enumerate}
    \item We present {\Dataset}, a novel dataset sourced from Reddit that captures multidimensional aspects of effective social support in mental health contexts, including posts, comments, feedback loops, and community-based metadata (e.g., upvotes, controversy scores) for nuanced evaluation of support quality.
    \item Building on social science and psychological theories, we propose a holistic framework for labeling effective social support, focusing on reciprocity and community reception, validated through human annotator evaluations, ensuring reliability, clarity, and real-world relevance.
    \item Our experiments show that leveraging {\Dataset} enhances LLMs' ability to produce context-sensitive, effective, and supportive responses.
\end{enumerate}


\section{Related Work}  
This section reviews related work across three key areas: social support datasets, LLMs for mental health support, and methods for measuring social support. 

\subsection{Social Support Datasets} 



The Emotion Support Conversation (ESConv) dataset \cite{liu2021towards}, is a foundational resource for emotional support dialogues. Despite its psychological comfort, its rule-based interactions limit real-world applicability. Medeiros et al. \cite{medeiros2018using} and Sharma et al. \cite{sharma2020computational} leveraged Twitter and Reddit data, respectively, to classify supportive interactions, offering insights into real-world scenarios. However, these datasets lack user feedback and focus narrowly on empathy or specific scenarios. Hosseini et al. \cite{hosseini2021takes} analyzed empathy in cancer support networks but focused on individual sentences in physical health contexts.

Our dataset addresses these gaps by incorporating diverse, real-world social media interactions, multiple support types, and user feedback to enable a comprehensive understanding of social support dynamics.
\subsection{Large Language Models for Mental Health Support}  
Large Language Models (LLMs) have shown promise in addressing mental health challenges through tasks like classification and summarization \cite{alghamdi2024less, xu2024mental}. Recent works such as Psy-LLM for psychological consultations \cite{lai2023psy}, ExTES for adaptive emotional responses \cite{zheng2023building}, SoulChat for empathetic dialogues \cite{chen2023soulchat}, and ChatCounselor for counseling \cite{liu2023chatcounselor} represent notable advancements. MindfulDiary \cite{kim2024mindfuldiary} offers journaling tools praised for emotional support. Despite these advances, challenges remain, including limited cultural diversity, overreliance on comforting language, and struggles with nuanced emotions \cite{zheng2023building, chen2023soulchat, liu2023chatcounselor}.

\subsection{Measuring Social Support} 
Evaluating social support on social media has evolved from indirect content analysis to mixed methods incorporating user feedback. Early work categorized comments by type and tone \cite{hale2019responding} or examined narrative features \cite{hale2020posting}. Later studies integrated sentiment analysis with engagement metrics \cite{raamkumar2020measuring} and analyzed comment content for empathy and guidance \cite{chen2021exploring}. These efforts often relied on indirect measures \cite{adelina2023stories}.

Recent advancements blend direct and indirect methods, such as surveys to track participant distress and health outcomes \cite{Zhou2021Changes, carter2023identity}, and regression analyses paired with quality-of-life metrics \cite{cahuas2023perceived}. Linguistic pattern analysis combined with user interactions \cite{morini2023can} highlights the importance of combining user feedback with quantitative metrics for comprehensive evaluation. This shift reflects the growing emphasis on mixed-method approaches to assess support effectiveness.



\begin{figure*}[h]

\centering
 \includegraphics[width=0.99\textwidth]{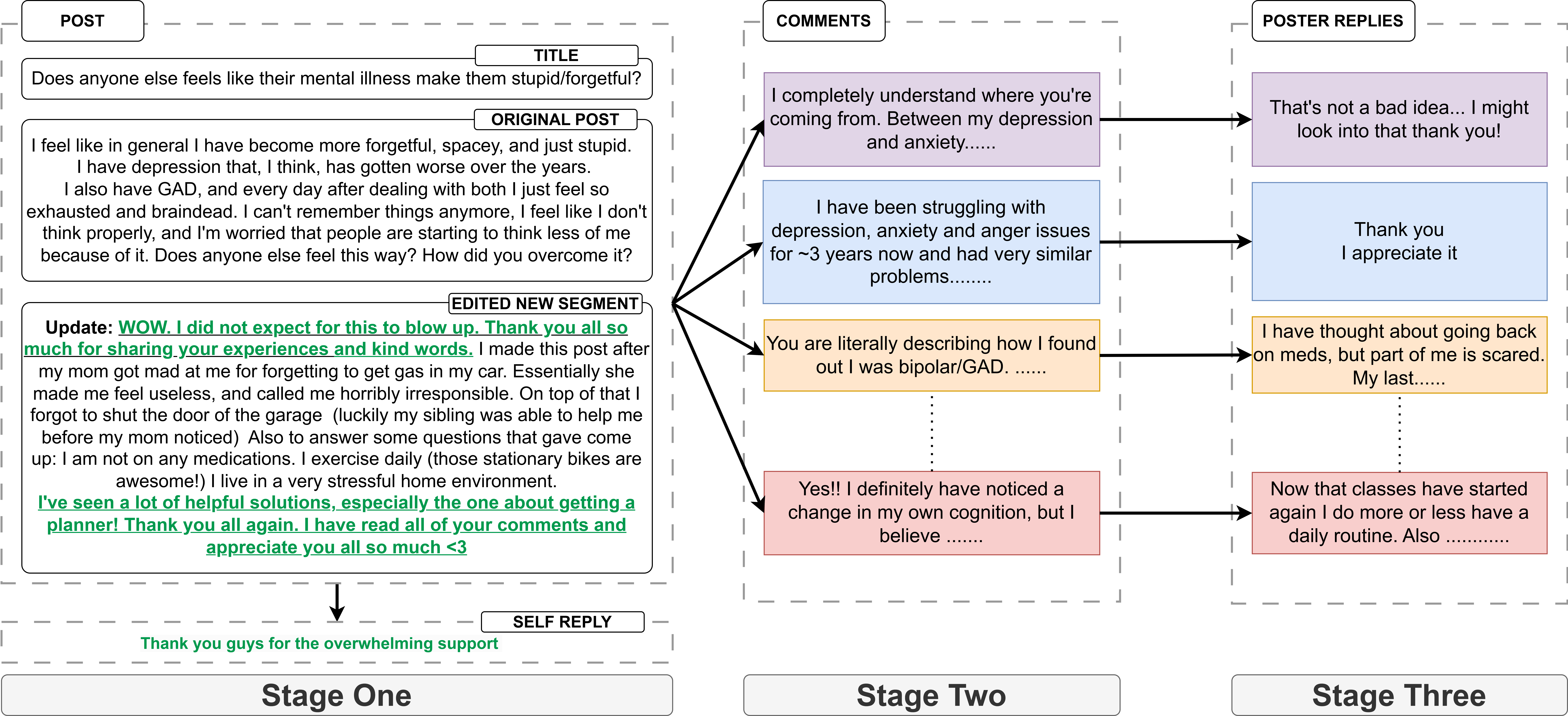}
  
  \caption{A real example of a mental health post from {\Dataset} showing three labeling stages.}
 \label{fig:example} 
\end{figure*}

\section{{\Dataset} Dataset}
This section presents a comprehensive overview of the dataset preparation process, outlining each step to ensure transparency and reproducibility. Specifically, we describe the methods used for data extraction, preprocessing, and labeling. A detailed description of the dataset contents and processing workflow is available in Appendix~\ref{app:dataset_page}.



\subsection{Data Extraction and Preprocessing }
To explore authentic expressions of mental health challenges and emotional venting, we focus on five key subreddit categories frequently analyzed in the literature~\cite{turcan2019dreaddit,rastogi2022stress}: post-traumatic stress disorder (PTSD), Depression, Anxiety, Stress, and general mental health. For data collection, we utilized the Python Reddit API Wrapper (PRAW)\footnote[2]{https://github.com/praw-dev/praw}.
We implemented various automatic mechanisms to filter out low-quality content, removing irrelevant posts such as spam, bots, advertisements, and surveys.
Among the remaining posts, we focused on those that met specific criteria: (1)~the post had been edited, (2)~it is related to mental health, (3)~it contains comments, (4)~the poster had responded to the comments, and (5)~neither the post nor the comments had been deleted.

The significance of an edited post lies in the indication that the user is actively reflecting on their content and the attention it has received.
To properly evaluate the social support received by the poster, we extracted comments along with auxiliary information, such as the date, likes, and comment controversy. This additional information helps provide a better perspective on the significance of each comment to the poster and the community. 
Next, we study the responses of the original poster to the comments, which offer insights into the interaction between the poster and the audience.
Additionally, we retained comments without replies from the original poster in the extended dataset for further analysis. These mental health-focused subreddit communities represent a diverse sample of over 2 million users seeking and offering social support. After cleaning and filtering, the final collection includes 59,666 comments linked to 1,689 unique posts. A golden subset, containing 8,507 comments with replies from the original poster, is associated with 1,098 unique posts. More details about this process are provided in Appendix~\ref{app:A1}.

\subsection{Data Labeling}
Our objective is to determine whether comments provide effective social support through a majority consensus derived from three human-centric annotation schemes: Social Support General Feedback Labeling, Social Support Engagement Labeling, and Social Support Individual Response Labeling. These schemes or stages were designed to capture perspectives from both the poster/user reciprocity and community reception. An illustration of the three stages with an example from {\Dataset} is shown in Figure~\ref{fig:example}.

\subsubsection{Stage One: Social Support General Feedback Labeling}
We employ complex regular expressions approaches to all posts to identify if the user perceived a social support. In most cases, users highlight this feedback either at the beginning or end of their post, often using phrases such as `Update' or `Edit'. Once extracted, we analyze the content of these edits to understand the user's motivation for making the changes.

We assign a label of 1 if the user explicitly reflects positively on the support received through comments. We assign a label of 0 when the user specifies the reason for the edit as simply updating the story, correcting grammar, or making unrelated changes without referencing the received support.

To further refine this labeling, we analyze all self-reply comments where the original poster responds to their own post. These self-replies are labeled using the same criteria as post edits.

Finally, for posts labeled as 1, all associated comments are categorized as supportive; for posts labeled as 0, the comments are categorized as non-supportive.
Our analysis revealed that 3,401 samples were labeled as 0, accounting for approximately 40\% of the dataset, while 5,102 samples were labeled as 1, representing about 60\%.

\subsubsection{Stage Two: Social Support Engagement Labeling}

In this labeling stage, we draw on the wisdom of the crowd. A comment is assigned a label of 0 if it receives dislikes, is marked as controversial, or has zero likes. This indicates that the community perceives the comment as unrelated, unworthy of support, or tension-inducing. Controversial comments have a relatively equal number of upvotes and downvotes, indicating a significant split in opinion on the topic within the community.
Notably, when a user posts a comment, it automatically receives one like. Therefore, we label it as 1 if the comment accumulates two or more likes, meaning at least one additional user found the comment helpful. We set this threshold low because comments may be buried by more popular ones or may receive lower engagement overall.

Our analysis revealed that 1,356 samples (approximately 16\%) received a label of 0, while 7,147 samples (approximately 84\%) were labeled as 1. To account for negative reception, we apply an overall multiplier of 0 for cases involving dislikes, zero likes, or controversy.

\subsubsection{Stage Three: Social Support Individual Response Labeling} 

This stage focuses on the individual's response to the comments, where the aim is to evaluate how the poster reflected on each comment. To determine whether a comment provides effective social support, we follow two key steps:
\begin{enumerate}
    \item Gratitude Detection: We utilize regular expressions to identify expressions of gratitude within the response.
    By capturing specific keywords indicative of gratitude, we assign a label of 1 if such expressions are present; otherwise, it is labeled as 0.
    \item Sentiment Analysis: We apply sentiment analysis~\cite{camacho-collados-etal-2022-tweetnlp} to assess the overall sentiment of the response. If the sentiment score exceeds a high threshold indicating the entire response is overwhelmingly positive, the response is labeled as 1; otherwise, it is labeled as 0.
\end{enumerate}

Further details are provided in Appendix \ref{app:stage_3}. The final stage three label, is derived as the product of gratitude detection and sentiment analysis labels.
Our analysis revealed that out of the total dataset, 4,576 samples (approximately 54\%) are labeled 0, while 3,927 samples (around 46\%) are labeled 1.


%


\subsection{Dataset Statistics}
The statistics of the data are presented in Table~\ref{tab:stat}. The Effective Social Support label (ESS) is the final aggregated label of the 3 stages and it includes 2,785 samples with a label of 0, comprising approximately 33\%, and 5,718 samples with a label of 1, accounting for around 67\%. The resulting golden dataset contains 1096 unique post ids and authors and 8503 unique comments with 6854 unique commenters. Please refer to Appendix~\ref{app:A2} for additional details.

\begin{table}[t]
\centering
\tiny
\setlength{\tabcolsep}{6.5pt}
\resizebox{\columnwidth}{!}{%
\begin{tabular}{l@{\hspace{4pt}}r@{\hspace{4pt}}r@{\hspace{4pt}}r@{\hspace{4pt}}r@{\hspace{4pt}}r@{\hspace{4pt}}r} 
\toprule
\multirow{2}{*}{\textbf{Subreddit}} & \multirow{2}{*}{\textbf{Posts}} & \multirow{2}{*}{\textbf{Pairs}} & \multicolumn{3}{c}{\textbf{Average Word Count}} & \multirow{2}{*}{\textbf{Label}}\\ 
\cmidrule(lr){4-6}
& &  & \textbf{Post} & \textbf{Comment} & \textbf{Reply} & \\ 
\midrule
\multirow{3}{*}{Anxiety} & \multirow{3}{*}{280} & 744 & 221 & 59 & 45 & 0 \\
 & & 1,881 & 186 & 51 & 28 & 1 \\
 & & 2,625 & 196 & 53 & 33 & All \\ 
\midrule
\multirow{3}{*}{Depression} & \multirow{3}{*}{261} & 795 & 223 & 61 & 42 & 0 \\
 & & 1,824 & 274 & 72 & 29 & 1 \\
 & & 2,619 & 258 & 69 & 33 & All \\ 
\midrule
\multirow{3}{*}{Mental Health} & \multirow{3}{*}{213} & 595 & 217 & 71 & 47 & 0 \\
 & & 1,194 & 199 & 55 & 27 & 1 \\
 & & 1,789 & 205 & 60 & 33 & All \\ 
\midrule
\multirow{3}{*}{PTSD} & \multirow{3}{*}{277} & 572 & 334 & 99 & 65 & 0 \\
 & & 780 & 295 & 103 & 38 & 1 \\
 & & 1,352 & 312 & 101 & 50 & All \\ 
\midrule
\multirow{3}{*}{Stress} & \multirow{3}{*}{65} & 79 & 289 & 70 & 74 & 0 \\
 & & 39 & 243 & 152 & 30 & 1 \\
 & & 118 & 274 & 97 & 60 & All \\ 
\midrule
\multirow{3}{*}{\textbf{Total}} & \multirow{3}{*}{1,096} & 2,785 & 257 & 72 & 55 & 0 \\
 & & 5,718 & 239 & 87 & 30 & 1 \\
 & & 8,503 & 249 & 76 & 42 & All \\
\bottomrule
\end{tabular}
}
\caption{Statistics of the golden dataset showing unique post counts and comment-reply pairs across mental health subreddits.}
\label{tab:stat}
\end{table}

\subsection{Human Annotation} 
To evaluate the performance of our labeling framework, we recruited subject matter experts to annotate the supportive content of multiple posts. Three annotators reviewed 564 human comments linked to 141 unique posts, representing approximately 13\% of the posts in the golden set. 

 The labeling framework aligned with human annotations for ESS=1 in 94.68\% of the cases, while agreement for ESS=0 was lower at 27.30\% . This difference reflects the strictness of our framework, which prioritizes identifying highly effective support, often underestimating supportiveness to reduce false positives. Comments labeled as ESS=0 frequently included cases that, while potentially supportive in broader human interpretations, did not meet the strict standards of reciprocity and validation defined by our labeling method.
\begin{figure}[h!]
    \centering
    \includegraphics[width=0.4\textwidth]{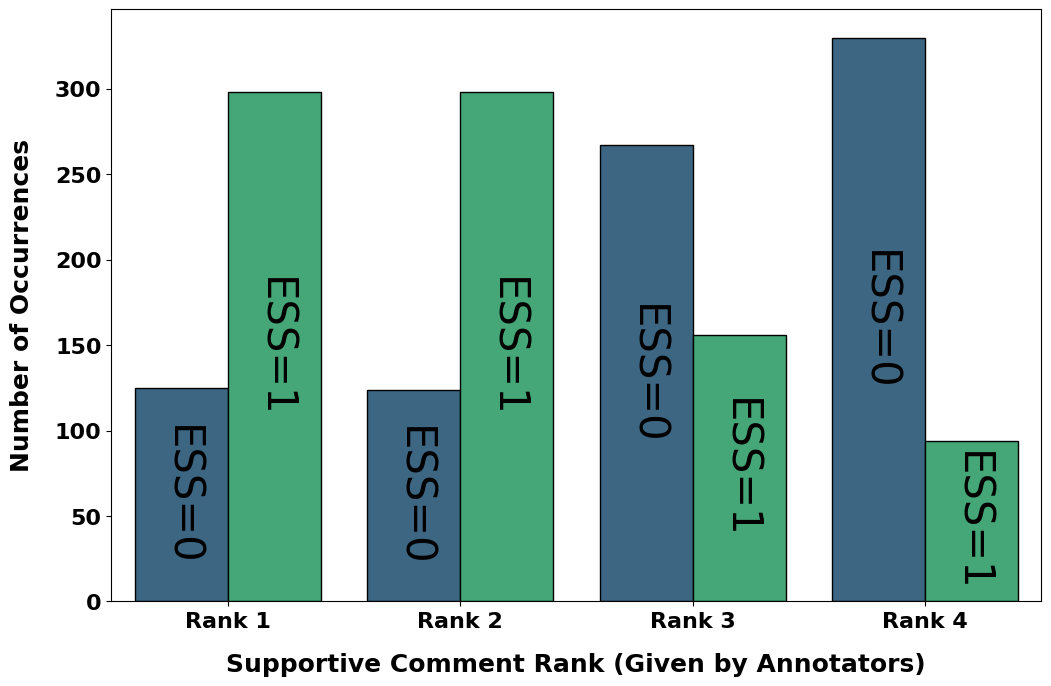} 
    \caption{Distribution of Annotator Rankings Across Effective and Non-Effective Social Support Labels.}
    \label{fig:ESS_rankings_fig}
\end{figure}
The annotators ranked comments from most to least supportive on a scale of 1 to 4, where rank 1 represents the most effective support and rank 4 the least. By analyzing how these rankings align with our Effective Social Support (ESS) label,as illustrated in figure \ref{fig:ESS_rankings_fig} we found that ESS = 1 is predominantly associated with ranks 1 and 2, while ESS = 0 is mostly linked to ranks 3 and 4. This confirms that our ESS label effectively differentiates between highly supportive and less or non-supportive comments. The ranking distribution was systematically validated to ensure accuracy, with total counts matching expected values based on annotator input. A detailed breakdown of this analysis can be found in Appendix ~\ref{app:ESS_rank}.

Cases labeled as ESS=1 by the majority showed stronger annotator agreement (75.42\% with full consensus) compared to ESS=0 cases (50\% with full consensus). Partial agreement (66.67\%) was common for ESS=0, occurring in 50\% of such cases. Overall, Fleiss’s Kappa on all labeled data is k=0.42, indicating `moderate agreement' ~\cite{fleiss1971measuring, landis1977measurement}.

\section{What Constitutes Effective Social Support?}

To investigate the factors that contribute to effective social support in a mental health context, we conducted a detailed analysis of our dataset. Our analysis is divided into two parts. First, we use Linguistic Inquiry and Word Count (LIWC) features to identify linguistic characteristics that distinguish supportive (ESS = 1) from non-supportive (ESS = 0) comments. Second, we categorize sub-types of ESS = 1 comments to explore the distribution and prevalence of different forms of effective support within our dataset.

\subsection{Linguistic Analysis of Comments}

LIWC captures psychological and social dimensions such as emotions, thinking styles, and social concerns~\cite{boyd2022development}. From the 118 linguistic features, we retain only those with a Pearson correlation of at least 0.1 and a p-value below 0.05 to identify key discriminative characteristics of effective support.

\noindent \textbf{Supportive vs. Non-Supportive Comments.} Supportive comments exhibited notable linguistic differences compared to non-supportive ones. They contained an average of \textit{37 additional positive words}, reflecting a more optimistic tone. A higher frequency of \textit{confidence-related (\textit{clout}) words} was also observed, averaging \textit{22 more such terms}, suggesting that supportive communicators often project authority and credibility. Social language, including politeness and communication-focused terms, was more prevalent in supportive comments, fostering a tone of \textit{empathy and engagement}. These comments also featured increased punctuation use, which further emphasized thoughtfulness. However, supportive comments were inversely associated with \textit{perceived authenticity}, possibly due to their polished and deliberate tone. 

In contrast, non-supportive comments were on average \textit{20 words longer} but lacked the positive sentiment and strategic language seen in supportive comments. This verbosity, instead of improving communication, often contributed to a \textit{negative emotional tone} and reduced perceived effectiveness.

\noindent \textbf{Replies to Supportive vs. Non-Supportive Comments.} Distinct linguistic and emotional differences emerged in replies. Responses to supportive comments were \textit{22 words longer} on average and demonstrated a \textit{more positive tone}, reflecting increased user engagement. These replies also exhibited a \textit{5\% increase in punctuation use}, indicative of greater thoughtfulness and emotional expression. On the other hand, replies to non-supportive comments tended to be shorter and often carried \textit{stronger negative emotional reactions}, highlighting the contrasting emotional dynamics triggered by supportive versus non-supportive interactions.

These findings underscore that effective supportive communication is characterized by being \textit{concise, positive, and authoritative}, with a rich use of \textit{social and empathetic language}. In contrast, non-supportive comments and their replies tend to lack these qualities, resulting in less engaging and emotionally negative interactions.

\subsection{Effective Support Categorization} 

Here, Building on House's~\cite{house1983work} framework of social support typology, we implemented a systematic classification of support patterns to analyze effective social support (ESS = 1) within our dataset. This classification scheme is structured around four primary dimensions of social support: i) \textit{Emotional Support}: Active listening, empathy expression, and validation of emotions. ii) \textit{Appraisal Support}: Affirmation, feedback, and social comparison. \textit{Informational Support}: Advice, guidance, and knowledge sharing. \textit{Instrumental Support}: Direct aid, practical assistance, and resource provision. This approach is informed by prior empirical validations~\cite{wortman1987conceptual,cohen1985stress,barrera1983structure,gottlieb1978development}, which underscore the unique contributions of each type of support. By categorizing comments according to these dimensions, we aim to systematically evaluate the prevalence and characteristics of effective support types. To this end, we employed GPT-4-turbo ~\cite{openai_gpt4_docs} as an LLM-annotator to analyze and identify the specific support type(s) associated with each comment.

\begin{figure}[h!]
    \centering
    \includegraphics[width=0.4\textwidth]{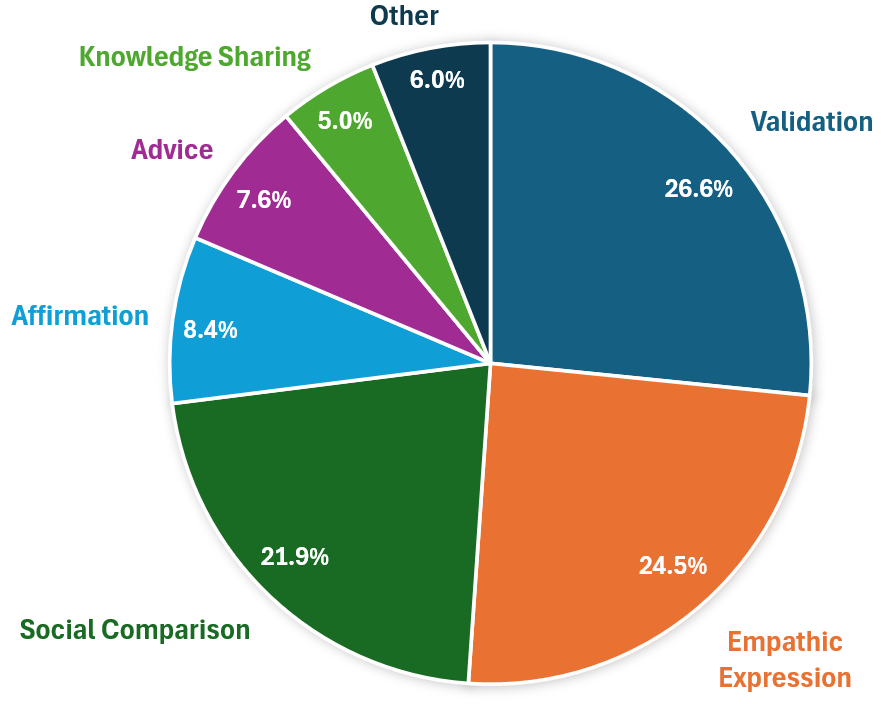} 
    \caption{Distribution of support types identified in the {\Dataset}. The chart shows the prevalence of validation, empathic expression, social comparison, affirmation, advice, and other categories as determined from the analysis.}
    \label{fig:support_types}
\end{figure}

As shown in Figure~\ref{fig:support_types}, validation, empathic expression, social comparison, affirmation, and advice emerged as the most common forms of support in our dataset. These patterns reflect the conversational and guidance-oriented nature of mental health discussions on Reddit, underscoring the diverse ways effective support is conveyed. The LLM-annotator was not formally evaluated as a classification tool; rather, it was used to assist analysis by generating support-type labels.

\begin{figure*}[t]
\centering
 \includegraphics[width=0.95\textwidth]
 {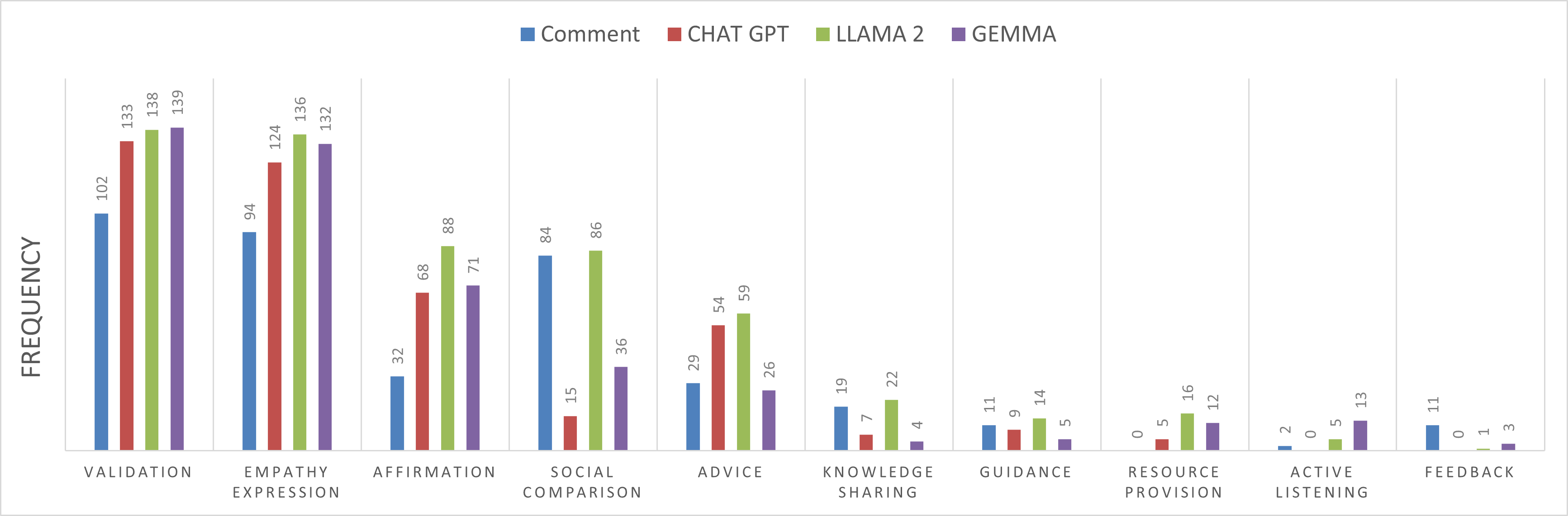}
  \caption{Support subcategory distribution across models showing comparative frequencies. Direct aid and Practical assistance appear less than 5 times across all models.
}
  \label{fig:supporttypes}
\end{figure*}

\section{LLM-Driven Effective Support}

In this section, we explore the potential of LLMs for generating effective social support in the context of mental health, leveraging {\Dataset}. First, we use the posts in {\Dataset} to evaluate the quality of social support generated by several LLMs, utilizing human annotators and the support categories introduced earlier. Next, we demonstrate how our dataset can be employed to improve LLMs’ ability to generate effective support through instruction tuning and direct preference optimization (DPO). Finally, we showcase the dataset's additional utility by training classification models to predict the effectiveness of social support comments, further solidifying its role in advancing both understanding and application of effective social support.

\subsection{Evaluating Effectiveness of Social Support Generated by LLMs}

We evaluate LLMs' ability to generate effective social support using posts from {\Dataset}. We selected the same subset of 141 posts (13\% of the golden set) used in the human annotations (Section 3) and queried three LLMs: Google's Gemma (7B)~\cite{team2024gemma}, Meta's Llama 2 (7B)~\cite{touvron2023llama}, and OpenAI's ChatGPT-3.5-turbo~\cite{openai_chatgpt_2023}. To ensure consistency, we fixed generation parameters (e.g., temperature = 0.8, see Appendix A.5). Annotators assessed the supportiveness of generated comments, judged whether human or LLM responses were more effective, and ranked the LLM outputs. The annotation process and bias considerations are detailed in Appendix~\ref{app:annotaion}. The evaluation revealed several notable trends: \\

\noindent \textbf{Preference for Human Comments:} Human comments were preferred in 49 cases (34.75\%), while LLMs were rated better in 21 cases (14.89\%). In 34 cases (24.11\%), support was rated equally. The remaining 37 cases (26.24\%) showed no agreement, reflecting the complexity and subjectivity of evaluating social support. \\

\noindent \textbf{LLM Performance Rankings:} Llama 2 consistently provided the highest-ranked support, demonstrating strength across diverse scenarios. ChatGPT showed competitive but variable performance, receiving both the highest and lowest ranks. Gemma was consistently the weakest but placed second in some cases. This variability reflects differences in model fine-tuning and alignment with human preferences.
More can be found in Appendix \ref{app:llms_rank}.

\noindent \textbf{Effective Support Categories in LLM Responses:}  
Emotional support was the most common type generated by all LLMs, as shown in Figure~\ref{fig:supporttypes}. LLaMA 2 excelled in producing emotional and appraisal support, outperforming other models in both frequency and quality. Human comments, however, showed fewer instances of validation and empathy but more often included affirmation and advice. Annotators noted that while LLMs mimicked supportive behaviors, their responses sometimes lacked authenticity, appearing overly dramatic or exaggerated. For instance, LLaMA 2’s social comparison attempts occasionally involved false claims, such as being a ``black man" or ``first responder" \cite{choi2023llms}. In contrast, human responses often shared relatable personal experiences, particularly in social comparison and knowledge-sharing contexts. These findings reveal key differences between human and LLM-generated support, highlighting both the strengths and limitations of LLMs. More details are provided in Appendix~\ref{app:A3}.

\subsection{LLM Alignment for Social Support}  

To enhance LLMs' capacity to generate effective social support, we aligned models using our dataset through a two-step process: supervised fine-tuning (SFT) and direct preference optimization (DPO). In SFT, models were trained on curated examples of effective support comments to develop a foundational understanding of empathy and validation. In DPO, pairwise comparisons of comments were used to distinguish the most and least effective responses. See Appendix~\ref{app:alignment} for data preparation details. Comments were classified as "chosen" (most effective) or "rejected" (least effective) based on label confidence, enabling the model to learn from human preferences and prioritize effective support qualities. Training parameters are detailed in Appendix~\ref{app:training}.

These alignment techniques significantly improved LLaMA’s performance. The aligned model outperformed the standard version, achieving a 71.6\% win-rate in human evaluations (i.e., in pairwise comparisons, aligned model responses were preferred 71.6\% of the time), as shown in Table~\ref{tab:llm_alignment_winrate}. This result underscores the potential of leveraging {\Dataset} to enhance LLMs for generating high-quality social support in mental health contexts.

\begin{table}[t]
\centering
\caption{\small Win-rate evaluation: LLaMA-2 aligned on RedditESS vs. original LLaMA-2.}
\vspace{-1mm}
\scriptsize 
\setlength{\tabcolsep}{3pt} 
\begin{tabular}{lc}
\toprule
\textbf{Comparison} & \textbf{Win-Rate (\%)} \\
\midrule
Aligned vs. Original & 71.6 \\
\bottomrule
\end{tabular}
\vspace{-3mm}
\label{tab:llm_alignment_winrate}
\end{table}

\begin{table}[t]
\centering
\caption{\small Support Identification Classification Performance.}
\vspace{-1mm}
\scriptsize 
\setlength{\tabcolsep}{3pt} 
\begin{tabular}{lccc}
\toprule
\textbf{Model} & \textbf{Accuracy (\%)} & \textbf{F1 (\%)} & \textbf{Avg Metric Score (\%)} \\
\midrule
BERT-base     & 75 & 84 & 80 \\
RoBERTa-base  & 76 & 85 & 83 \\
\bottomrule
\end{tabular}
\vspace{-3mm}
\label{tab:classification_metrics}
\end{table}


\subsection{Classification of Effective Social Support}

In addition to alignment efforts, we used our dataset to train classification models capable of predicting whether a given comment provides effective social support. This task aimed to build models that could evaluate the supportiveness of a comment based on its linguistic and contextual features. To achieve this, the dataset was divided into 90\% training and 10\% testing splits, ensuring no overlap between posts in the training and testing sets. This careful division prevented the models from memorizing specific posts and ensured robust generalization. More details on data pre-processing steps are provided in Appendix~\ref{app:A4}.

As summarized in Table~\ref{tab:llm_alignment_winrate} and Table~\ref{tab:classification_metrics}, we evaluated the performance of PLM models fine-tuned on this task, as well as the effectiveness of LLaMA-2 aligned using RedditESS. The BERT-based model achieved an accuracy of 75\%, an F1-score of 84\% (measured on the positive class, i.e., effective social support), and a combined evaluation score of 80\% (average of accuracy, precision, recall, and F1-score). RoBERTa-base demonstrated superior performance, achieving an accuracy of 76\%, an F1-score of 85\% (positive class), and a combined evaluation score of 83\%. These results underline the utility of our dataset for developing classifiers capable of identifying effective social support comments. Furthermore, they demonstrate the potential of transformer-based architectures for addressing nuanced tasks such as evaluating the quality of social support.

\section{Conclusion and Future Work}

The increasing role of AI systems in providing mental health support necessitates a deeper understanding of what makes such support truly effective. Through {\Dataset}, we advance this understanding by providing a comprehensive dataset that captures the multifaceted nature of effective social support in real-world digital mental health communities. By incorporating user feedback loops, and community validation metrics, our dataset moves beyond the traditional emphasis on empathy alone to encompass the broader spectrum of support mechanisms valued by individuals seeking help.
Our rigorous evaluation process, combining human annotation with automated analysis, demonstrates the dataset's reliability and practical utility. The successful integration of {\Dataset} into LLM training pipelines shows promising results in enhancing AI systems' ability to provide more nuanced, context-aware support. These improvements suggest that AI-driven mental health support systems can be developed to better reflect the complexity and diversity of human support needs.


\section*{Limitations}
While {\Dataset} provides valuable insights into social support interactions, it has several limitations. The dataset focuses on posts where authors actively engaged by editing their content, which, while offering a unique perspective, restricts the dataset size and completeness. Edited or removed content can obscure the context, including the emotional tone and specific issues raised, complicating the analysis of how effectively comments address the original concerns.  

Additionally, our study does not explicitly account for the intentions behind posts or comments, such as whether users are seeking or offering support. This lack of intention analysis may lead to misclassifications, such as labeling critical or neutral comments as supportive based on surface features. Incorporating intention recognition could enhance the alignment of classifications with user intent.  
We acknowledge that Reddit’s voting system does not fully capture real-world interpersonal support; however, the use of upvotes forms only one of three labeling stages, reducing its overall influence on the final support label. This approach draws on previous studies suggesting that controversy, upvotes, or likes can correlate with users’ perception of social approval or support. Although our dataset centers on a single platform, we have selected multiple subreddits with diverse user bases and time spans, as outlined in the appendix, to approximate a broad range of interactions. We recognize that using Reddit alone is an inherent limitation, but this multi-stage labeling strategy and the breadth of communities included help mitigate bias concerns.

Our multi-stage labeling approach introduces potential biases. For example, using a threshold of two likes to identify effective support may overlook detailed responses with fewer likes while prioritizing less substantive comments that meet the threshold. Similarly, discrepancies between human annotations and automated labels highlight challenges in capturing gratitude or perceived support effectiveness, particularly when these signals conflict with like-based thresholds.

Our allignment experiment was designed to demonstrate the dataset's utility in improving LLM performance for support-related tasks. Due to limitations in human annotator and resource budgets, we focused on comparing the best ranking "LLAMA-2" base LLM responses with aligned LLM responses, and did not include comparisons with human-generated comments or evaluate multiple LLMs.

A label imbalance in the dataset, with a predominance of supportive comments, may bias models toward overestimating the effectiveness of support. Balancing the dataset through resampling or threshold adjustments could mitigate this issue.  

Finally, the limited number of unique posts associated with multiple comments may introduce learning biases in pre-trained language models (PLMs), leading to an overreliance on post-specific features and reduced generalizability. Further work is needed to disentangle post-specific and comment-specific features to enhance model robustness and applicability across diverse contexts.

\section*{Ethical Statement}

In this study, we developed a dataset, referred to as {\Dataset}, containing real mental health interactions sourced from publicly available Reddit posts and comments. We acknowledge the sensitive nature of mental health-related data and have taken comprehensive steps to prioritize ethical considerations, user privacy, and data security throughout the research process.  
\begin{itemize}
    \item \textbf{Data Filtering and Privacy Protection:} Posts and comments deleted by the posters and commenters' as of January 2024 were excluded from the dataset. All personally identifiable information (PII), including usernames, was replaced with placeholders such as `[USER]`. URLs were replaced with `[LINK]`, and subreddit names were replaced with `[SUBREDDIT]` to further anonymize the data.  
   \item \textbf{Publicly Available Data Usage:} This dataset was constructed exclusively from publicly accessible data, and no private or non-consensual sources were used. While Reddit's terms of service permit the use of public data for research, we acknowledge the ethical implications of working with sensitive content and have made every effort to minimize harm.  
   \item \textbf{Minimizing Harm and Avoiding Stigmatization:} We recognize that mental health content can be deeply personal and may unintentionally cause distress if misused. Thus, we emphasize that {\Dataset} is intended solely for research purposes aimed at improving mental health support systems. It should not be used for commercial exploitation or any application that could stigmatize or harm individuals.
   \item \textbf{Annotation Ethics:} The annotation process was conducted with a focus on respecting the context and intent of the original posters. Annotators were trained to handle the content sensitively and instructed to approach the task with empathy, avoiding biases or harmful judgments.
   \item \textbf{LLM Alignment for Ethical Applications:} In line with our research goals, we ensure that any LLM alignment using {\Dataset} aims to improve the quality of context-sensitive and genuinely supportive responses. The aligned models are not intended to replace professional mental health services but rather to complement them by offering preliminary support or guidance.
   \item \textbf{Compliance with Ethical Guidelines:} The research protocol was reviewed to ensure compliance with ethical guidelines for working with social media data. Any future use of this dataset should similarly adhere to relevant ethical standards and data protection laws. 
\end{itemize}


\bibliography{paper}

\appendix
\label{sec:appendix}
\section{Appendix}

\subsection{ More Details on Dataset Curation}
~\label{app:A1}
Collectively, these mental health-centered subreddit communities encompass over 2 million subscribers, offering a diverse sample of individuals seeking and providing social support.

We aided our data with relevant scraped data based on the Post IDs form "Dreaddit"\cite{turcan2019dreaddit} which allowed us to bypass API scraping challenges and access historical post data.

The entire dataset after cleaning and filtering contains 59.666 comments/samples and it associated with 
1,689 unique post ids, the golden set is the subset with replies from the original poster and it consists of 8,507 comments with replies and is associated with 1,098 unique post IDs. Moreover, the silverset is the dataset without the replies have 51,159 comments and is associated with 1,514 unique post id, and there are 923 post IDs that are common across the golden and silverset , those were the posts that had some comments with replies and some did not receive original poster reply. understanding why some social support comments have received a response from the poster while some did not is crucial for future studies.

\textbf{For an overview of the distribution of posts over the years, see Table~5 for details.}
\begin{table}[h!]
\centering
\small
\label{tab:unique_post_ids_comparison}
\begin{tabular}{|c|c|c|c|}
\hline
Year & Combined & Goldset & Silverset \\ \hline
2010 & 1 & 1 & 0 \\ \hline
2011 & 2 & 1 & 2 \\ \hline
2012 & 9 & 5 & 8 \\ \hline
2013 & 23 & 12 & 19 \\ \hline
2014 & 17 & 9 & 13 \\ \hline
2015 & 26 & 14 & 19 \\ \hline
2016 & 31 & 15 & 27 \\ \hline
2017 & 85 & 46 & 68 \\ \hline
2018 & 128 & 81 & 109 \\ \hline
2019 & 219 & 137 & 214 \\ \hline
2020 & 390 & 258 & 376 \\ \hline
2021 & 272 & 189 & 263 \\ \hline
2022 & 124 & 87 & 108 \\ \hline
2023 & 176 & 128 & 151 \\ \hline
2024 & 186 & 115 & 137 \\ \hline
\textbf{Total} & \textbf{1,689} & \textbf{1,098} & \textbf{1,514} \\ \hline
\end{tabular}
\caption{Unique post IDs comparison across datasets, the combined dataset that contains the silver and golden set.}
\end{table}

\subsection{ Further Dataset Analysis}
~\label{app:A2}

\textbf{Agreement of the Three stages labels:} It is worth mentioning that for "stage one" it is labeled 1 if demonstrates unexpected happiness (e.g. expressed via likes, number of comments, or awards), or otherwise acknowledges the value of the social support. conversely labeled 0, if no specific reason for the edit is provided, we assume that insufficient social support was received and label it as 0.

We have found that we have 523 samples with 0 labels in all metrics which is around 6\% and we have 2464 samples with 1 labels across all metrics which which is around 29\% , and we had only one label 1 for 2248 samples which is around 27\% ,and we had only two label 1 for 3268 samples which is around 38\%.

\textbf{Dataset Temporal Analysis:} To further examine community dynamics, we analyzed the temporal aspects of social support, including the time of post creation, edits, comments, and replies. While different subreddits exhibit small variations in timing, several consistent patterns emerge. On average, users edit their posts to acknowledge receiving meaningful support approximately 10.8 days (259.14 hours) after the original post’s creation. This delay reflects thoughtful engagement and a willingness to provide feedback or updates. Supportive comments are typically posted 2.95 days (70.75 hours) after the post's creation, highlighting the community’s promptness in offering assistance. Furthermore, users take an average of 1.33 days (31.98 hours) to reply to these comments, demonstrating timely acknowledgment and appreciation of the support received.

These temporal trends underscore the evolving nature of posts, as users interact with supportive comments and provide updates. They highlight the critical role of community responses in fostering meaningful participation and facilitating emotional and practical support.

\textbf{Dataset Activity Analysis:}The analysis reveals notable insights into subreddit activity. On average, each post on the Anxiety subreddit receives 9.38 comments, with a total of 2,625 unique comments distributed in 280 unique posts. Similarly, depression exhibits a high level of engagement, averaging 10.03 comments per post, with 2,619 unique comments across 261 unique posts. Mental health follows with 8.40 comments per post, 1,789 unique comments, and 213 unique posts. In contrast, PTSD and stress show lower activity levels, with 4.88 and 1.82 average comments per post, respectively. PTSD has 1,352 unique comments on 277 unique posts, while Stress contains 118 unique comments on 65 unique posts. These figures illustrate varying levels of user engagement and content distribution across subreddits. We also notice that on average the PTSD subreddit contains the longest posts with an average word count of 312 on the other end of the spectrum. Anxiety subreddit contained the lowest average word count of 196 words.

\textbf{Dataset Flairs Analysis:}The dataset contains user-input flairs, which are tags applied by posters to indicate the type of request associated with their posts. In our curated golden dataset, 46\% of the samples are included, which encompass a variety of 48 unique link-flair texts (as observed in Table 5). The analysis of these flairs reveals key patterns in the types of support sought within the community.

Emotional support emerges as the most prevalent category, reflected in flairs such as `Venting', `Needs A Hug/Support', and `Sadness/Grief'. These flairs indicate posts where users share vulnerabilities and seek empathy, validation, or comfort. Informational and practical support represents another significant category, with flairs like `Advice', `Question', and `Health', where users request specific guidance, experiential insights, or actionable steps, often pertaining to personal struggles or mental health.

Flairs such as `Progress!' and `Success!' highlight positive reinforcement, encouraging celebratory responses and fostering motivation through the recognition of milestones and achievements. Additionally, community engagement flairs, including `DAE Questions and Discussion', facilitate shared experiences and intellectual dialogue, thereby promoting a sense of belonging. Specialized support needs are also evident, with flairs like `Trigger Warning (TW)' addressing sensitive topics with care and `Help A Loved One' signaling indirect support for others. These findings underscore the diverse and multifaceted dynamics of support within the community, ranging from emotional validation and problem solving to celebratory and reflective interactions.

\begin{table*}[ht]
\centering
\small
\begin{tabular}{|l|r|p{10cm}|}
\hline
\textbf{Category} & \textbf{Percentage} & \textbf{Link Flair Text} \\
\hline
Support and Advice & 23.41\% & Uplifting, Inspiration / Encouragement, Share Your Victories, Helpful Tips! \\
Mental State and Emotions & 23.21\% & Mental Health, Emotional Support \\
Questions and Discussion & 21.43\% & Questions, Discussions \\
Progress and Positive News & 16.67\% & Helpful, Share Your Victories \\
Uncategorized & 6.94\% & Work/School, Driving, Safe mode: voting off, friend, Off My Chest, Resources, aftermath, Relationships, Work/Search, School/Exams, Help A Loved One, Lifestyle, Subreddit Challenge, Family/Relationship, Meta, Relationship \\
Health and Treatment & 4.17\% & Health, Treatment \\
TW (Trigger Warning) & 4.17\% & Trigger Warning \\
\hline
\end{tabular}
\caption{Percentages of each category in our golden set, categories were best match}
\label{table:flair_categories}
\end{table*}

\subsection{Annotation Procedure and Bias Considerations}
~\label{app:annotaion}
Human annotators were tasked with assessing the effectiveness of social support provided in comments. The procedure involved two phases:
\begin{itemize}
    \item \textbf{Supportiveness Annotation:} Annotators evaluated whether each comment provided effective social support based on predefined criteria. Comments were labeled as ESS=1 (supportive) if they demonstrated reciprocity and validation within the community, or ESS=0 (non-supportive) otherwise.
    \item \textbf{Ranking of Comments per Post:} For each post, annotators compared all associated comments and assigned rankings based on their relative supportiveness. Rank 1 indicated the most supportive comment, rank 2 the second-best, and lower ranks (3, 4, etc.) represented less supportive or non-supportive comments. This ensured fine-grained assessment of effectiveness within the context of each post.
    \item \textbf{LLM vs. Human Judgments:} In the evaluation of LLM-generated comments, annotators were informed that comments originated from either humans or LLMs (Gemma, LLaMA 2, ChatGPT), but were not told which specific comments belonged to each group, nor were they aware of prior ESS labels. This approach was designed to reduce bias and collect unbiased comparative judgments on which comments offered the most effective support. Annotators assessed whether human or LLM-generated comments were preferred and ranked the effectiveness of LLM-generated responses across the models.
\end{itemize}

\noindent \textbf{Bias Considerations:} While annotators knew that both human and LLM-generated comments were included, the blind evaluation of individual comments helped minimize bias. However, we acknowledge that subtle stylistic differences between human and LLM-generated content may have inadvertently influenced judgments.

\noindent \textbf{Additional Details:} \begin{itemize} \item \textbf{Anonymity and Randomization:} Annotators were not informed which comments were human-written or LLM-generated, nor were they shown prior ESS labels. To reduce bias, comments were randomly ordered, and the source identity was hidden. 
\item \textbf{Win-Rate Evaluation Task:} For a separate “win-rate” evaluation, annotators compared responses from the standard LLaMA-2 model and the aligned LLaMA-2 model on the same post, without knowing which model produced each response. \item \textbf{Prior Exposure Consideration:} We acknowledge that some annotators had previously encountered LLM-generated content, but randomization, limiting context, and source anonymity were employed to mitigate potential bias. \end{itemize}

\subsection{Stage 3 Extended Information }
\label{app:stage_3}

In Stage 3, we label a reply as supportive if it includes common expressions of gratitude (e.g., “thank you,” “thanks,” “I appreciate”). To capture expressions of gratitude. We encode these variations as regular expressions (e.g., detecting common word stems like “thank” or “grateful” and handling punctuation/case variations) to ensure broad coverage.  This method is designed to capture genuine acceptance of support while reducing the risk of false positives from borderline or sarcastic or insincere acknowledgments.

Explicit gratitude expressions serve as well-established markers of supportive interactions, as validated by previous research \cite{ chen2024enhancing,islam2024leveraging,ho2023pilot,sciara2021gratitude,yoshida2022network}. These studies validate the role of gratitude expressions in fostering social bonds and prosocial behavior, supporting our keyword-based method for detecting genuine acceptance of support.

Moreover, a high sentiment threshold of 0.75 (on a scale of 0 to 1) is highly recommended in established research demonstrating that stringent thresholds in sentiment classification enhance precision and minimize false positives 
\cite{li2025semantic,liu2024self}. Findings from these studies underscore that higher thresholds effectively eliminate ambiguous or borderline classifications, refining the precision and reliability of sentiment analysis models. This threshold ensures that only responses with a demonstrably strong positive sentiment are classified as supportive, enhancing the robustness and accuracy of our classification model.

More importantly, Stage 3 is just one component of a multi-stage process. It does not directly dictate the final label; that requires a majority consensus across all stages. In this way, we reduce the likelihood that any potential misclassification at a single stage will compromise the overall reliability of our labeling approach.

\subsection{Further Human Observations Findings}
~\label{app:A3}
The annotations and comparisons among three annotators reveal key patterns in supportive commentary, directly highlighting differences between human and LLM-generated responses. Human-generated responses provide context-aware support, using personal narratives and insights to reflect genuine empathy. For instance, a single word like "hugs" can be meaningful to a poster, demonstrating how minimal yet personally resonant support from humans can gain high engagement. On the other hand, human support can also falter—some replies become harsh, self-centered, or dismissive. There are cases where humans respond to a poster's negative venting ~\cite{alghamdi2023studying} with a "pity party" scenario, offering negative camaraderie that was nonetheless deemed effective by the annotators' labeling stages because it aligned with what the poster wanted to hear. At the same time, human commenters can be more creative, providing innovative perspectives and nuanced solutions drawn from their lived experiences rather than a general template.

In contrast, LLM-generated support, such as from ChatGPT, Gemma, and Llama, tends to be more uniform and sometimes overly dramatic or patronizing. Annotators noted that LLM outputs vary in length and style—ChatGPT produces shorter responses, while Llama generates longer texts with emoji usage (sometimes inappropriately in serious situations). Gemma tends to over-interpret emotions and comment on writing style. LLMs frequently rely on repetitive phrases like "sending you love" or "I'm here for you," which feel forced and fail to deeply engage with the poster's issues. Although Llama produces longer messages, this additional length often translates into rambling, generic reassurance rather than deeper understanding. They also sometimes suggest unrealistic outcomes (e.g., no pain after surgery) or provide resources that may not align with the poster's context~\cite{agrawal2024can,agrawal2024mindful}. Still, LLMs excel in consistently acknowledging posters' difficulties and encouraging help-seeking behaviors, even if such encouragement is formulaic and lacks personalized insight.

Recent research supports these observations.\cite{lee2024large} found that LLM-generated messages were consistently rated as more empathetic than human-written ones, although their uniformity sometimes lacked the variability found in human responses. Similarly, \cite{welivita2024large} noted that LLM-generated support, while empathetic, often exhibited a consistent style that might feel impersonal in complex situations. The challenges LLMs face in navigating complex emotional and cultural contexts , emphasizing their limitations in achieving genuine emotional understanding. \cite{havaldar2023multilingual} further demonstrated that multilingual LLMs often reflect Western norms, even when responding in other languages, indicating a lack of cultural nuance. Similarly, \cite{shen2024understanding} found significant discrepancies in LLMs' grasp of cultural commonsense, highlighting inherent biases in their understanding. Research by \cite{li2023large} explored LLMs' emotional intelligence, revealing their limited capacity to fully comprehend and respond to emotional stimuli. \cite{amirizaniani2024llms} evaluated LLMs' Theory of Mind reasoning, noting their struggles with achieving human-like social reasoning in open-ended responses.

Some human comments are judged as supportive even when they simply commiserate with negative sentiments, because that's what the original poster desired. Conversely, some LLM-sounding human responses—vague, patronizing, or detached—are labeled as not supportive. Annotators noted that human support can be messy, including the occasional use of aggressive language or cursing, yet still be seen as relatable or effective due to its authenticity. LLMs, lacking genuine personal investment, often fail to achieve this resonance. They may try to mimic human experiences (``As a black person myself..." or ``I've been there too") in an attempt to relate, but these attempts sometimes ring hollow without the deeper context and sincerity that a human can bring.

In summary, while LLM responses are consistent and reliably encouraging, they often lack the emotional and contextual richness that human supporters provide. Humans can craft messages that draw on personal struggles and shared understanding to offer practical and empathetic solutions. They excel at finding positive aspects within difficult situations while maintaining honesty about challenges. Even brief human replies or seemingly offbeat responses can resonate deeply if they align with the poster's emotional needs. LLMs perform adequately in simple, straightforward cases but struggle with complex situations requiring multiple layers of understanding or community context (like past post history or community-specific knowledge). The data suggest that true supportive engagement thrives on authenticity, contextual awareness, and sincerity—traits inherently more accessible to human commenters than to LLMs.

\subsection{PLM Models Preprocesisng}
~\label{app:A4}
For preprocessing, post-comment pairs were tokenized up to 512 tokens, with priority given to the entire length of the comment. Any remaining token space was filled with content from the associated post. This approach ensured that the model captured the full context of the comment while maintaining relevance to the post. Additionally, comments associated with the same post ID were kept exclusively within either the training or testing set to avoid data leakage.

\subsection{Large Language Models Support Generation Prompt Design}
~\label{app:A5}
In our methodology, we developed a standardized prompt to elicit supportive responses from Large Language Models (LLMs) when analyzing social media posts. The core prompt was structured as:
\textit{"The following is a Reddit post posted by a social media user; then, provide a supportive comment for their post: [POST]"}
This prompt design emphasizes directness and clarity to ensure consistent interpretation across different LLM architectures. We specifically chose the terminology "supportive comment" to guide models toward generating emotionally aware responses while maintaining sufficient flexibility to examine how different LLMs naturally interpret and execute supportive behavior. The prompt uniformity across all tested models was essential for ensuring valid cross-model comparisons and reproducibility of results.

\subsection{LLM Alignment dataset}
\label{app:alignment}
To improve how Large Language Models (LLMs) generate effective social support that better mirrors high-quality human support patterns, we developed a comprehensive training approach using a substantial dataset. We selected approximately 55\% (33,000 samples) of our combined dataset through random sampling to ensure representative coverage while maintaining computational efficiency.
Our methodology leverages our previously validated RoBERTa-based social support classification model, which has demonstrated strong performance in identifying supportive content. To create a sophisticated ranking system for posts and their associated comments, we developed a multi-dimensional scoring framework that incorporates three key metrics:
First, we utilize the probability scores from our pre-trained language model (PLM), which provides an initial assessment of the supportive nature of each comment. Second, we conduct sentiment analysis\cite{camacho-collados-etal-2022-tweetnlp} on the comments, calculating the probability of supportive content using the same approach established in stage three of our original labeling process. Third, we compute the percentile rank of each comment's likes relative to other comments on the same post, normalizing this engagement metric within the context of each discussion.
Each of these three measurements produces a value between 0 and 1, which we then sum to create a composite score ranging from 0 to 3. To account for potentially problematic content, we apply a negative multiplier (-1) to comments tagged with either "dislike" or "controversy" flags. This adjustment helps ensure that controversial or potentially harmful content receives appropriate weighting in our ranking system. Finally, we sort the comments based on these adjusted scores in descending order, with higher scores indicating more effective and well-received supportive comments.
This refined approach allows us to systematically identify and rank supportive comments while accounting for both content quality and community reception. The resulting ranked dataset provides a strong foundation for training LLMs to generate more effective social support that aligns with successful human support patterns.

\subsection{Effective Social Support Human Ranking }
\label{app:ESS_rank}

In total, 564 comments were manually annotated, and 472 of those were labeled as “supportive” (label 1) by human annotators. Our aggregated labeling approach marked 282 comments as label 1, of which 267 overlapped with the human annotations (267/282 = 94.7\%).
Moreover,out of the annotated comments, 472 were labeled as ESS=1 (supportive), and 92 as ESS=0 (non-supportive). The labeling framework aligned with human annotations for ESS=1 in 94.68\% (447/472) of cases, while agreement for ESS=0 was lower at 27.30\% (25/92).

We tasked the annotators with ranking each comment based on the level of supportiveness, from best support (most effective) to worst support (least effective). The ranking system follows a 1 to 4 scale, where rank 1 represents the most effective support and rank 4 represents the least supportive or least effective comment. Since the annotators initially labeled comments only as "supportive" or "not supportive" in a binary manner, this ranking system allows us to further differentiate the degree of supportiveness. By analyzing how these rankings align with our Effective Social Support (ESS) label, we can determine whether a supportive comment is also highly effective and qualifies as ESS = 1, or if it is supportive but not effective enough to be considered ESS = 1 and instead falls under ESS = 0, or if it is entirely non-supportive (ESS = 0).

In Figure \ref{fig:ESS_rankings_fig}, we present the distribution of rankings across the ESS label. The results indicate that most comments classified as ESS = 1 (effective support) were ranked as rank 1 or rank 2, suggesting a strong correlation between ESS = 1 and high supportiveness. Conversely, most comments classified as ESS = 0 were assigned rank 3 or rank 4, indicating that when a comment does not qualify as ESS = 1, it is perceived as offering either limited support or no meaningful support at all. This analysis reinforces that our ESS label successfully captures the highest level of supportiveness, distinguishing between highly effective support and comments that are supportive but not truly effective or entirely lacking support.

To ensure accuracy in our analysis, we calculated the frequency of each ranking category (1, 2, 3, 4) and how often they were assigned to comments labeled as effective social support (ESS = 1) or non-effective support (ESS = 0). Since each comment received rankings from three annotators, the total count of rankings should equal three times the number of unique comments in the dataset. By restructuring the data using the melt function, we transformed the separate ranking columns from different annotators into a single column, allowing us to count occurrences systematically. We then grouped the data by rank and ESS label to determine how frequently each ranking was associated with effective or non-effective support. The final count was validated against the expected total rankings, ensuring no missing values or discrepancies. The results confirm a strong alignment between lower ranking numbers (1 and 2) and effective support, while higher rankings (3 and 4) are more frequently assigned to non-effective support, reinforcing the reliability of both the ranking system and the ESS label.

\begin{table}[t]
\centering
\small

\resizebox{\columnwidth}{!}{%
\begin{tabular}{lccc}
\toprule
\textbf{Condition} & \textbf{AN1} & \textbf{AN2} & \textbf{AN3} \\
\midrule
Rank 1 or 2, ESS=1 & 269 (47.70\%) & 272 (48.23\%) & 274 (48.58\%) \\
Rank 3 or 4, ESS=1 & 126 (22.34\%) & 187 (33.16\%) & 218 (38.65\%) \\
Rank 1 or 2, ESS=0 & 13 (2.30\%) & 8 (1.42\%) & 6 (1.06\%) \\
Rank 3 or 4, ESS=0 & 156 (27.66\%) & 95 (16.84\%) & 65 (11.52\%) \\
\bottomrule
\end{tabular}
}
\caption{Supportive Comment Ranking by Annotators (AN1, AN2, AN3), from best (1) to worst (4).}
\label{tab:supportive_ranking}
\end{table}

Recognizing that supportive comments vary in effectiveness, annotators also provided fine-grained annotations by identifying the most supportive comment for each post, followed by the second-best, and so on. For each post, comments were annotated with rank 1 for the most supportive, rank 2 for the next best, and lower ranks (3, 4) for less supportive or non-supportive comments. As shown in Table~\ref{tab:supportive_ranking}, comments labeled ESS=1 were frequently marked as the top (rank 1) or second-best (rank 2) support, whereas ESS=0 comments were rarely assigned to these top positions.

\subsection{LLMs Ranking Extended }
\label{app:llms_rank}
\begin{figure}[h!]
    \centering
    \includegraphics[width=0.4\textwidth]{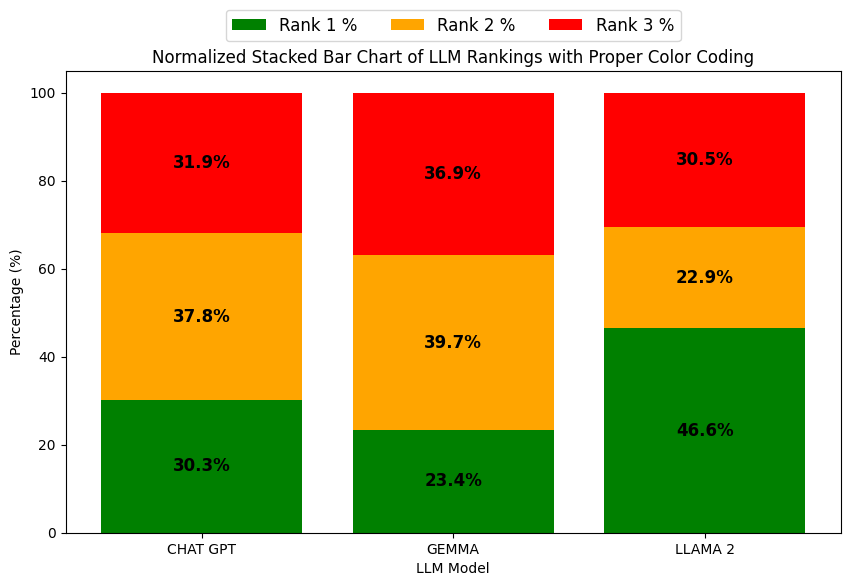} 
    \caption{Comparative Performance of LLaMA 2, ChatGPT, and Gemma Based on Human Annotator Rankings}
    \label{fig:LLM_RANKSSSS}
\end{figure}

Figure \ref{fig:LLM_RANKSSSS} illustrates the distribution of Rank 1 (green), Rank 2 (yellow) , and Rank 3 (red)  assignments for each LLM, highlighting LLaMA 2's dominance as the most preferred model, ChatGPT’s balanced performance, and Gemma’s lower ranking. The color-coded visualization provides a clear comparison of how frequently each model was rated as the best, middle, or least supportive option.

The ranking analysis shows that LLaMA 2 (LLM3) was the most preferred model, receiving the highest percentage of Rank 1 assignments (46.57\%), indicating that it consistently provided the best support in evaluations. ChatGPT (LLM1) had a more balanced performance, with Rank 2 being the most frequent, suggesting it was often a reliable choice but not consistently the top performer. On the other hand, Gemma (LLM2) was the least preferred, earning the lowest percentage of Rank 1 assignments (23.40\%) and the highest percentage of Rank 3 (36.88\%), meaning it was frequently considered the weakest option. This ranking highlights LLaMA 2’s strong performance, ChatGPT’s solid but middle-ground positioning, and Gemma’s struggles, offering valuable insights into their relative effectiveness and potential areas for improvement.

\subsection{Training Parameters for LLM Alignment}
\label{app:training}

\begin{table}[H]
\centering
\caption{Training and Hyperparameter Configurations for SFT, DPO, and LoRA}
\begin{tabular}{ll}
\toprule
\multicolumn{2}{c}{\textbf{Supervised Fine-Tuning (SFT)}} \\
\midrule
Model              & Meta's LLaMA-2-7b-chat-hf      \\
Hardware           & Nvidia A100 (40GiB)            \\
Optimizer          & Fused AdamW                    \\
Learning Rate      & 2e-4                           \\
Precision          & Mixed precision (bf16)         \\
Epochs             & 3                              \\
Gradient Clipping  & 0.3                            \\
Warmup             & 3\% of total steps             \\
Max Sequence Length & 1024 tokens                    \\
\midrule
\multicolumn{2}{c}{\textbf{Direct Preference Optimization (DPO)}} \\
\midrule
Beta               & 0.1 (divergence control)       \\
Loss Function      & Sigmoid-based DPO loss         \\
Batch Size         & 4 per device (effective: 4)    \\
Learning Rate      & 5e-5                           \\
Epochs             & 3                              \\
Max Prompt Length  & 910 tokens (95th percentile)   \\
Max Sequence Length & 2060 tokens                    \\
\midrule
\multicolumn{2}{c}{\textbf{LoRA Configurations (SFT \& DPO)}} \\
\midrule
LoRA Alpha         & 128                            \\
Dropout            & 0.05                           \\
Rank               & 256                            \\
Target Modules     & "all-linear"                   \\
\bottomrule
\end{tabular}
\end{table}

\section{Repository Insights: Structure, Data, and Key Information}
\label{app:dataset_page}
This appendix provides a structured breakdown of the files available in our GitHub repository and the key insights that can be extracted from them. These datasets and resources are designed to facilitate replication, further analysis, and innovation in research.

\subsection{Repository Access}
Our repository is accessible at:  
\url{https://anonymous.4open.science/r/RedditESS-3577}  

\subsection{Dataset Files and Their Contents}

\subsubsection{Extended\_liwc\_features\_goldset.zip}  
This file contains the gold set data along with all Linguistic Inquiry and Word Count (LIWC) features and relevant metadata, linked to anonymized keys. It enables researchers to replicate our feature extraction process, validate insights, and extend analyses beyond the scope of this study.

\subsubsection{Goldset\_with\_aggregated\_final\_label.zip}  
This dataset includes all anonymized keys along with their corresponding labels and extracted three-stage values. It provides essential insights into the distribution of labels and extracted values across the gold set.

\subsubsection{LLMs\_Social\_Support\_Classes\_golden.zip}  
This file contains the large language model (LLM)-generated support responses for all posts in the gold set. It allows researchers to examine variations in how different LLMs generate supportive responses, extract additional linguistic and contextual features, and uncover novel insights. Additionally, it provides a foundation for replicating our findings and assessing LLM annotation performance.

\subsubsection{RedditESS\_Combined\_dataset\_with\_anonymized\_columns.zip}  
This is the most comprehensive dataset, comprising both the gold and silver sets, with unique anonymized keys. It is instrumental in analyzing differences between comments that received a reply from the original poster and those that did not, contributing to research on engagement in online support interactions.

\subsubsection{RedditESS\_Silverset.zip}  
This subset of the combined dataset excludes the gold set. It consists of comments that did not receive a reply from the original poster, providing a valuable contrast for engagement analysis.

\subsubsection{Social\_Support\_Classes\_Comments\_Goldset.zip}  
This file contains all gold set comments along with ChatGPT-4’s classification of social support categories. It enables researchers to replicate our classification methodology, analyze support categories, and build upon our findings.

\subsubsection{Concluding Remarks}
By providing these datasets, we aim to support the research community in replicating our results, extending the study of online social support, and fostering new avenues for exploration.

\end{document}